\documentclass[aos,noinfoline]{imsart}
\usepackage{fullpage}
\usepackage{times}
\usepackage{amsmath,amssymb}
\usepackage{natbib}
\usepackage[caption=false]{subfig}
\usepackage{graphicx}
\usepackage{url}
\usepackage{booktabs,siunitx}
\usepackage[labelformat=simple,justification=raggedright,format=hang,font={small,it},singlelinecheck=false]{caption}
\usepackage{macros}
\usepackage{floatrow}
\usepackage{titlesec}

\setattribute{journal}{name}{}

\titleformat{\section}[hang]
  {\bfseries}{\thesection.}{1em}{}

\begin{document} 

\begin{frontmatter}
\runtitle{Input Warping for Bayesian Optimization}
\title{Input Warping for Bayesian Optimization of Non-stationary Functions}

\begin{aug}
  \author{\fnms{Jasper}  \snm{Snoek}\ead[label=e1]{jsnoek@seas.harvard.edu}},
  \author{\fnms{Kevin} \snm{Swersky}\ead[label=e2]{kswersky@cs.toronto.edu}},
  \author{\fnms{Richard S.}  \snm{Zemel}\ead[label=e3]{zemel@cs.toronto.edu}}
  \and
  \author{\fnms{Ryan P.}  \snm{Adams}\ead[label=e4]{rpa@seas.harvard.edu}}

  \runauthor{Snoek et al.}

  \affiliation{Harvard University and University of Toronto}

  \address{Jasper Snoek\\
   		School of Engineering and Applied Sciences\\
        Harvard University\\
        \printead{e1}}

  \address{Kevin Swersky\\
   		Department of Computer Science\\
        University of Toronto\\
        \printead{e2}}

\address{Ryan P. Adams\\
   		School of Engineering and Applied Sciences\\
        Harvard University\\
        \printead{e4}}

  \address{Richard S. Zemel\\
   		Department of Computer Science\\
        University of Toronto\\
        \printead{e3}}
  
\end{aug}

\newcommand{\fix}{\marginpar{FIX}}
\newcommand{\new}{\marginpar{NEW}}
\makeatletter
\newcommand{\pushright}[1]{\ifmeasuring@#1\else\omit\hfill$\displaystyle#1$\fi\ignorespaces}
\newcommand{\pushleft}[1]{\ifmeasuring@#1\else\omit$\displaystyle#1$\hfill\fi\ignorespaces}
\makeatother

\begin{abstract}
Bayesian optimization has proven to be a highly effective methodology for the global optimization of unknown, expensive and multimodal functions.  The ability to accurately model distributions over functions is critical to the effectiveness of Bayesian optimization.  Although Gaussian processes provide a flexible prior over functions, there are various classes of functions that remain difficult to model.  One of the most frequently occurring of these is the class of non-stationary functions.  The optimization of the hyperparameters of machine learning algorithms is a problem domain in which parameters are often manually transformed \emph{a priori}, for example by optimizing in ``log-space,'' to mitigate the effects of spatially-varying length scale.  We develop a methodology for automatically learning a wide family of bijective transformations or \emph{warpings} of the input space using the Beta cumulative distribution function.  We further extend the warping framework to multi-task Bayesian optimization so that multiple tasks can be warped into a jointly stationary space. On a set of challenging benchmark optimization tasks, we observe that the inclusion of warping greatly improves on the state-of-the-art, producing better results faster and more reliably.
\end{abstract}
\end{frontmatter}

\section{Introduction}

Bayesian optimization is a strategy for the global optimization of noisy, black-box functions.  The goal is to find the minimum of an expensive function of interest as quickly as possible.  Bayesian optimization fits a surrogate model that estimates the expensive function, and a proxy optimization is performed on this in order to select promising locations to query.  Naturally, the ability of the surrogate to accurately model the underlying function is crucial to the success of the optimization routine.  Recent work in machine learning has revisited the idea of Bayesian optimization~\citep[e.g.,][]{osborne-2009a,Brochu2010,Srinivas2010,hutter-2011a,BergstraJ2011,Bull2011,snoek-etal-2012b,hennig-schuler-2012} in large part due to advances in the ability to efficiently and accurately model statistical distributions over large classes of real-world functions.  Gaussian processes (GPs)~\citep[see, e.g.,][]{Rasmussen2006} provide a powerful framework to express flexible prior distributions over smooth functions, yielding accurate estimates of the expected value of the function at any given input, but crucially also uncertainty estimates over that value.  These are the two main components that enable the exploration and exploitation tradeoff that makes Bayesian optimization so effective.

A major limitation of the most commonly used form of Gaussian process regression is the assumption of stationarity --- that the covariance between two outputs is invariant to translations in input space. This assumption simplifies the regression task, but hurts the ability of the Gaussian process to model more realistic non-stationary functions. This presents a challenge for Bayesian optimization, as many problems of interest are inherently non-stationary.  For example, when optimizing the hyperparameters of a machine learning algorithm, we might expect the objective function to have a short length scale near the optimum, but have a long length scale far away from the optimum.  That is, we would expect bad hyperparameters to yield similar bad performance everywhere (e.g., classifying at random) but expect the generalization performance to be sensitive to small tweaks in good hyperparameter regimes.

We introduce a simple solution that allows Gaussian processes to model a large variety of non-stationary functions that is particularly well suited to Bayesian optimization.  We automatically learn a bijective \emph{warping} of the inputs that removes major non-stationary effects.  This is achieved by projecting each dimension of the input through the cumulative distribution function of the Beta distribution, while marginalizing over the shape of the warping.  Our approach is computationally efficient, captures a variety of desirable transformations, such as logarithmic, exponential, sigmoidal, etc., and is easily interpretable.  In the context of Bayesian optimization, understanding the parameter space is often just as important as achieving the best possible result and our approach lends itself to a straightforward analysis of the non-stationarities in a given problem domain.  

We extend this idea to multi-task Bayesian optimization \citep{swersky-etal-2013a} so that multiple tasks can be warped into a jointly stationary space.  Thus, tasks can be warped onto one another in order to better take advantage of their shared structure.

In the empirical study that forms the experimental part of this paper, we show that modeling non-stationarity is extremely important and yields significant empirical improvements in the performance of Bayesian optimization.  For example, we show that on a recently introduced Bayesian optimization benchmark~\citep{Eggensperger-etal-2013a}, our method outperforms all of the previous state-of-the-art algorithms on the problems with continuous-valued parameters.  We further observe that on four different challenging machine learning optimization tasks our method outperforms that of~\citet{snoek-etal-2012b}, consistently converging to a better result in fewer function evaluations.  As our methodology involves a transformation of the inputs, this strategy generalizes to a wide variety of models and algorithms.  Empirically, modeling non-stationarity is a fundamentally important component of effective Bayesian optimization. 

\section{Background and Related Work}
\subsection{Gaussian Processes}
The Gaussian process is a powerful and flexible prior distribution over functions~${f : \mathcal{X}
  \rightarrow \mathbb{R}}$ which is widely used for non-linear Bayesian regression.  An attractive property of the Gaussian process in the context of Bayesian optimization is that, conditioned on a set of observations, the expected output value and corresponding uncertainty of any unobserved input is easily computed.

The properties of the Gaussian process are specified by a mean function~${m: \mcX \to \reals}$ and a positive definite covariance, or kernel,
function~${K: \mcX \times \mcX \to \reals}$.  Given a finite set of training points~$\mathrm{I}_N = \{\brmx_n, y_n\}^N_{n=1}$, where ${\brmx_n\in\mcX},\ {y_n \in \mathbb{R}}$, the predictive mean and
covariance under a GP can be respectively expressed as:
\small
\begin{align}
\mu(\brmx; \mathrm{I}_N) &= m(\brmX) + K(\brmX,\brmx)^{\top}K(\brmX,\brmX)^{-1} (\brmy - m(\brmX)), \\
\Sigma(\brmx,\brmx' ; \mathrm{I}_N) &= K(\brmx,\brmx') - K(\brmX,\brmx)^{\top} K(\brmX,\brmX)^{-1} K(\brmX,\brmx').
\end{align}
\normalsize
Here~$K(\brmX,\brmx)$ is the~$N$-dimensional column vector of
cross-covariances between~$\brmx$ and the set~$\brmX$.  The~${N \times N}$ matrix
$K(\brmX,\brmX)$ is the Gram matrix for the set~$\brmX$ resulting from applying the covariance function $K(\brmx,\brmx')$ pairwise over the set $\{{\brmx_n}\}_{n=1}^N$.
The most common choices of covariance functions~$K(\brmx,\brmx')$ are functions of~${r(\brmx,\brmx') = \brmx - \brmx'}$, such as the automatic relevance determination (ARD) exponentiated quadratic covariance 
\begin{flalign}
K_{SE}(\brmx, \brmx') = \theta_0 \exp(-r^2) &\quad r = \sum^D_{d=1}(x_d-x'_d)^2/\theta_d^2\;,
\end{flalign}
or the ARD Mat\'{e}rn~$5/2$ kernel advocated for hyperparameter tuning with Bayesian optimization by \citet{snoek-etal-2012b}:
\begin{align}
  K_{\sf{M52}}(\brmx,\brmx') = \theta_0
  \left(
  1+ \sqrt{5 r^2} + \frac{5}{3}r^2
  \right)
  \exp\left\{-\sqrt{5 r^2}\right\}.
\end{align}

Such covariance functions are invariant to translations along the input space and thus are \emph{stationary}.  

\subsection{Non-stationary Gaussian Process Regression}
Numerous approaches have been proposed to extend GPs to model non-stationary functions.  \citet{gramacy-2005} proposed a Bayesian treed GP model which accommodates various complex non-stationarities through modeling the data using multiple GPs with different covariances.  Various non-stationary covariance functions have been proposed~\citep[e.g.,][]{higdon-etal-98a,Rasmussen2006}.  Previously, \citet{sampson-guttorp-92a} proposed projecting the inputs into a stationary latent space using a combination of metric multidimensional scaling and thin plate splines.  \citet{schmidt-ohagan} extended this warping approach for general GP regression problems using a flexible GP mapping.  Spatial deformations of two dimensional inputs have been studied extensively in the spatial statistics literature~\citep{anderes-2008a}.  \citet{bornn-etal-2012a} project the inputs into a higher dimensional stationary latent representation.  \citet{snelson-etal-2003} apply a warping to the output space,~$\brmy$, while \citet{adams-stegle-2008a} perform input-dependent output scaling with a second Gaussian process.

Compared to these approaches, our approach is relatively simple, yet as we will demonstrate, flexible enough to capture a wide variety of nonstationary behaviours. Our principal aim is to show that addressing nonstationarity is a critical component of effective Bayesian optimization, and that any advantages gained from using our approach would likely generalize to more elaborate techniques.

\subsection{Multi-Task Gaussian Processes}
Many problems involve making predictions over multiple datasets (we will henceforth refer to these prediction problems as tasks). When the datasets share an input domain, and the mappings from inputs to outputs are correlated, then these correlations can be used to share information between different tasks and improve predictive performance. There have been many extensions of Gaussian processes to the multi-task setting, e.g.,~\citet{goovaerts1997geostatistics,alvarez2011computationally}. However, a basic and surprisingly effective approach is to assume that each task is derived from a single latent function which is transformed to produce each output~\citep{teh-etal-2005a, bonilla-etal-2008a}.

Formally, this approach involves combining a kernel over inputs $K(\brmx,\brmx')$ and a kernel over task indices $K(t,t')$, $t=\left\{1,...,T \right\}$ via a product to form the joint kernel:
\begin{align}
K((\brmx,t),(\brmx',t')) &= K_T(t,t')K(\brmx,\brmx').
\end{align}
We infer the elements of $K_T(t,t')$ directly using the spherical parametrization of a covariance matrix~\citep{osborne-thesis,pinheiro-1996}.

\subsection{Bayesian Optimization}
Bayesian optimization is a general framework for the global
optimization of noisy, expensive, black-box
functions~\citep{Mockus1978}, see \citet{Brochu2010} or \citet{lizotte-thesis} for an in-depth explanation and review.  The strategy relies on the use of a relatively cheap probabilistic model that can be queried liberally as a surrogate in order to more effectively evaluate an expensive function of interest.  Bayes' rule is used to derive the posterior estimate of the true function, given observations, and the surrogate is then used to determine, via a proxy optimization over an \emph{acquisition function}, the next most promising point to query.  Using the posterior mean and variance of the probabilistic model, the acquisition function generally expresses a tradeoff between exploitation and exploration.  Numerous acquisition functions and combinations thereof have been proposed~\citep[e.g.,][]{kushner-1964a,Srinivas2010,hoffman-etal-2011}.

In this work, we follow the common approach, which is to use a GP to define a
distribution over objective functions from the input space to a loss that
one wishes to minimize.  Our approach is based on that of~\citet{Jones2001}. Specifically, we use a GP surrogate, and the \emph{expected improvement} acquisition function~\citep{Mockus1978}.
Let~${\sigma^2(\brmx) = \Sigma(\brmx,\brmx)}$ be the marginal predictive variance of a GP, and define
\begin{align}
\gamma(\brmx) &= \frac{f(\brmx_\text{best}) - \mu(\brmx; \left\{\brmx_n, y_n\right\}, \theta)}{\sigma(\brmx; \left\{\brmx_n,y_n\right\},\theta)}\;,
\end{align}
where $f(\brmx_\text{best})$ is the lowest observed value. The expected improvement criterion is defined as
\begin{align}
a_{EI}(\brmx; \left\{\brmx_n,y_n\right\},\theta) &= \sigma(\brmx; \left\{\brmx_n,y_n\right\},\theta) \left (\gamma(\brmx) \Phi(\gamma(\brmx)) \right. \nonumber \\
&\quad \left. +\, \mathcal{N}(\gamma(\brmx); 0,1)\right ).
\end{align}
Here $\Phi(\cdot)$ is the cumulative distribution function of a standard normal, and~$\mathcal{N}(\cdot; 0,1)$ is the density of a standard normal. Note that the method proposed in this paper is independent of the choice of acquisition function and do not affect its analytic properties.

\subsection{Multi-Task Bayesian Optimization}
When utilizing machine learning in practice, a single model will often need to be trained on multiple datasets. This can happen when e.g., new data is collected and a model must be retrained. In these scenarios we can think of each dataset as a different task and use multi-task Gaussian processes to predict where to query next. In \citet{krause-ong-2011}, this idea was applied to find peptide sequences that bind to molecules for vaccine design, while in \citet{swersky-etal-2013a} it was applied to hyperparameter optimization. In these cases it was shown that sharing information between tasks can be extremely beneficial for Bayesian optimization. Other approaches include \citet{remi2013collab}, which finds a joint latent function over tasks explicitly using a ranking model, and \citet{hutter-2011a} which uses a set of auxiliary task features to improve prediction.

\begin{figure*}[t]
\centering%
  \includegraphics[clip=true,trim=0 50 0 0,width=0.85\textwidth]{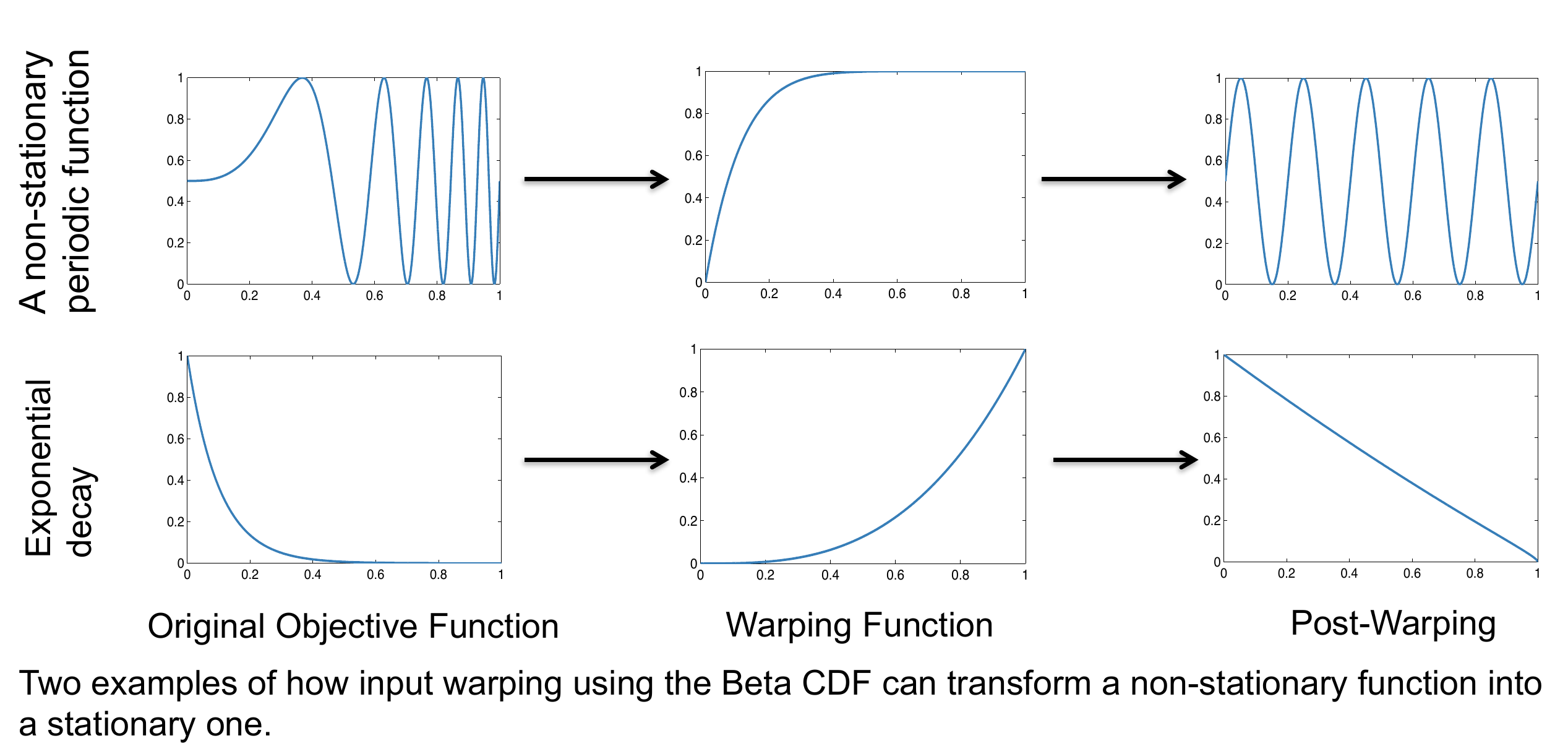}
\caption{Two examples of how input warping using the Beta CDF can transform a non-stationary function into a stationary one.  The warping function maps the original inputs on the horizontal axis to new inputs shown on the vertical axis.  The effect is to stretch and contract regions of the input space in such a manner as to remove non-stationarity.}
\label{fig:warpedfunctions}
\end{figure*}

\section{Input Warping}
We assume that we have a positive definite covariance function $K(\brmx, \brmxhat)$, where $\brmx, \brmxhat \in [0,1]^D$ due to projecting a bounded input range to the unit hypercube.  In practice, when tuning the hyperparameters of an algorithm, e.g., the regularization parameter of a support vector machine, researchers often first transform the input space using a monotonic function such as the natural logarithm and then perform a grid search in this transformed space.  Such an optimization in ``log-space" takes advantage of \emph{a priori} knowledge of the non-stationarity that is inherent in the input space.  Often however, the non-stationary properties of the input space are not known \emph{a priori} and such a transformation is generally a crude approximation to the ideal (unknown) transformation.  Our approach is to instead consider a class of bijective warping functions, and estimate them from previous objective function evaluations.  We can then use commonly-engineered transformations---such as the log transform---to specify a prior on bijections.  Specifically, we change the kernel function to be $K(w(\brmx), w(\brmxhat))$,
\begin{align}
w_d(\brmx_d) &= \mathrm{BetaCDF}(\brmx_d;\alpha_d,\beta_d) \, , \nonumber \\
&= \int_{0}^{\brmx_d} \frac{u^{\alpha_d-1} (1-u)^{\beta_d-1}}{B(\alpha_d,\beta_d)}\mathrm{d}u \, ,
\end{align}
where BetaCDF refers to the Beta cumulative distribution function and $B(\alpha,\beta)$ is the normalization constant. That is, ${w : [0,1]^D \rightarrow [0,1]^D}$ is a vector-valued function in which the~$d$th output dimension is a function of the~$d$th input dimension, and is specified by the cumulative distribution function of the Beta distribution.  Each of these~$D$ bijective transformations from~$[0,1]$ to~$[0,1]$ has a unique shape, determined by parameters~${\alpha_d > 0}$ and~${\beta_d > 0}$. The Beta CDF has no closed form solution for non-integer values of $\alpha$ and $\beta$, however accurate approximations are implemented in many statistical software packages.

Alternatively, one can think of input warping as applying a particular kind of non-stationary kernel to the original data. Examples of non-stationary functions and their corresponding ideal warping that transforms them into stationary functions are shown in Figure~\ref{fig:warpedfunctions}.

Our choice of the Beta distribution is motivated by the fact that it is capable of expressing a variety of monotonic warpings, while still being concisely parameterized. In general, there are many other suitable choices.

\begin{figure*}[t]
\centering%
\subfloat[Linear\label{fig:linear_beta}]{%
  \includegraphics[width=0.24\textwidth]{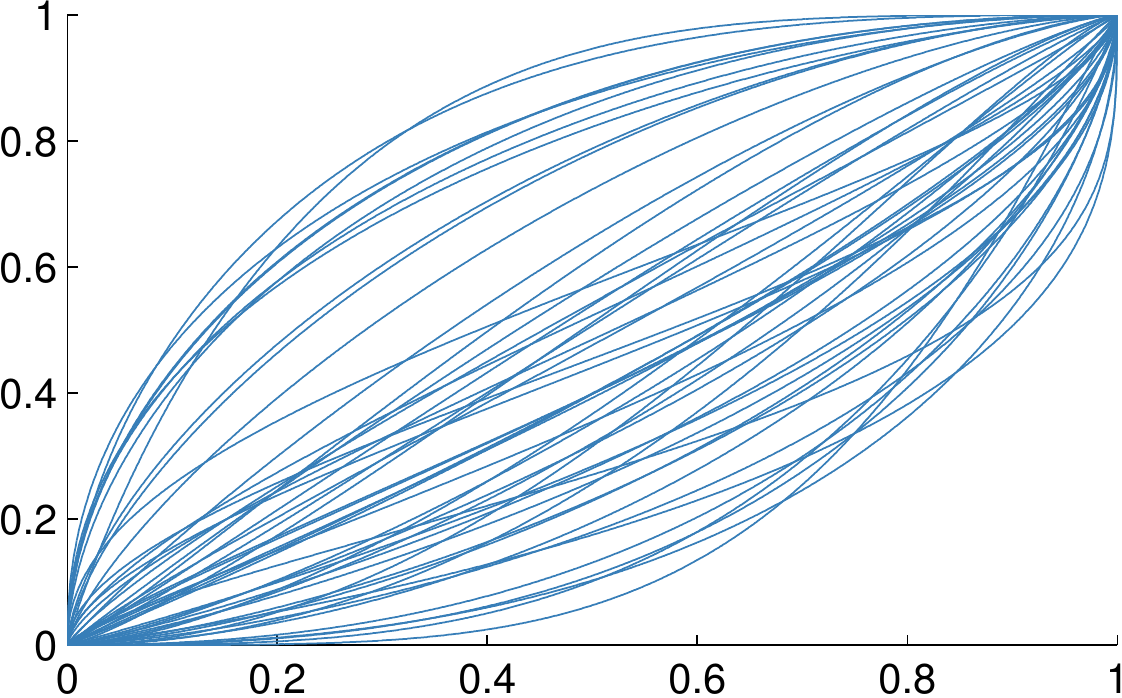}}
\subfloat[Exponential\label{fig:exp_beta}]{%
  \includegraphics[width=0.24\textwidth]{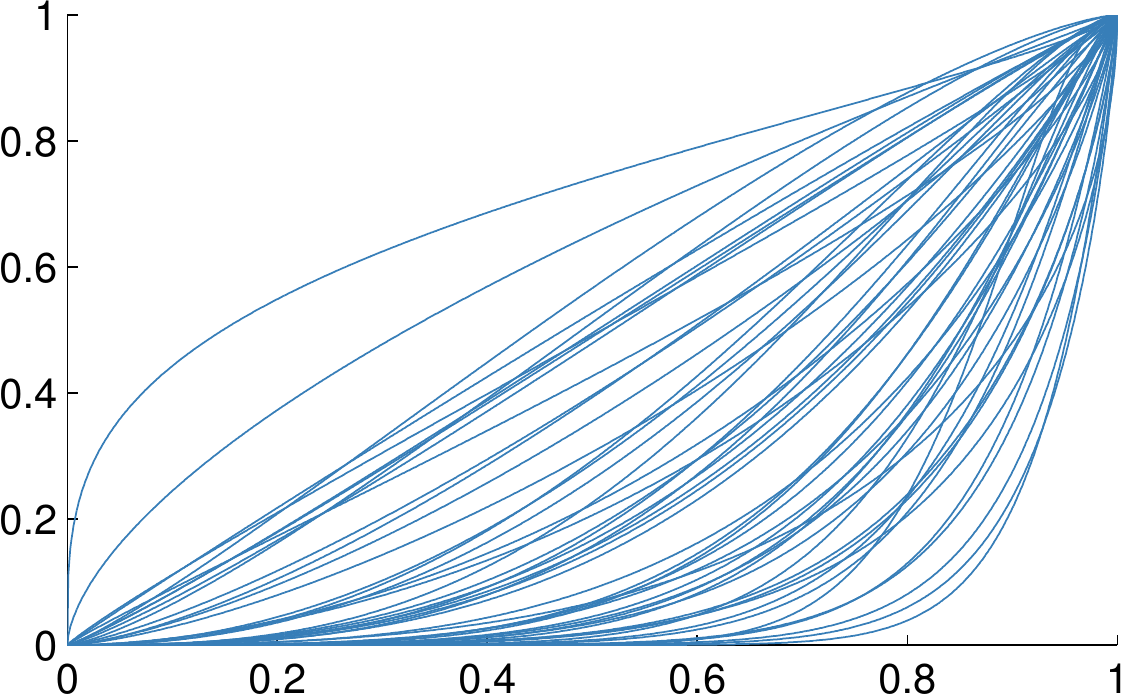}}
  \subfloat[Logarithmic\label{fig:log_beta}]{%
  \includegraphics[width=0.24\textwidth]{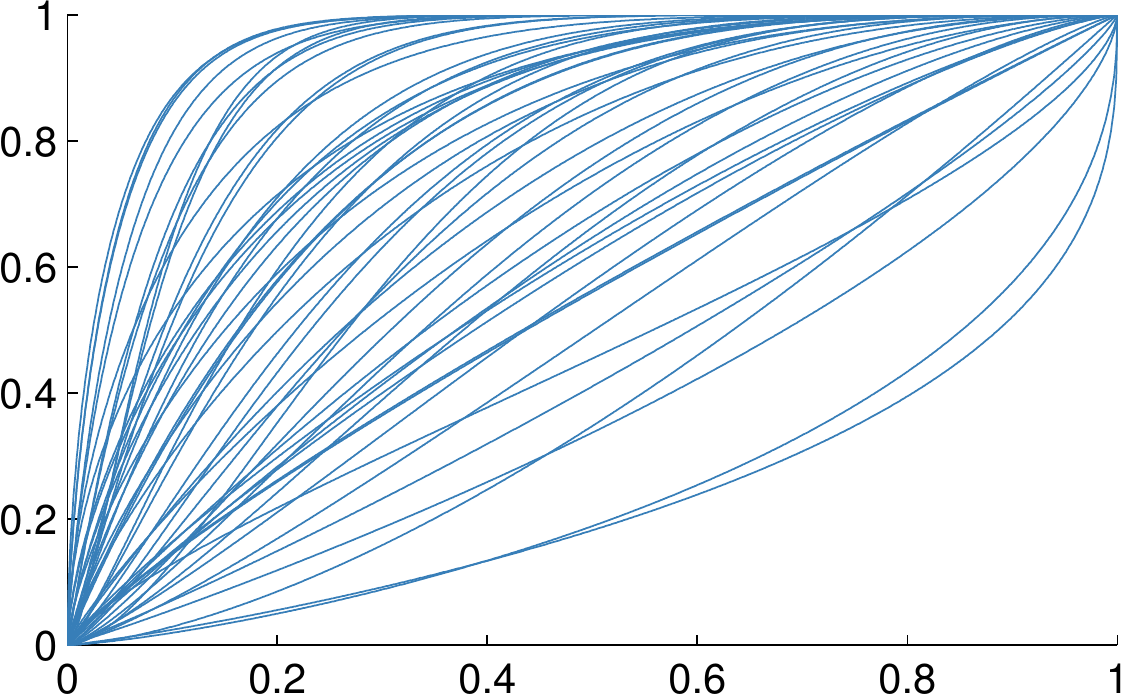}}
    \subfloat[Sigmoidal\label{fig:sigmoidal_beta}]{%
  \includegraphics[width=0.24\textwidth]{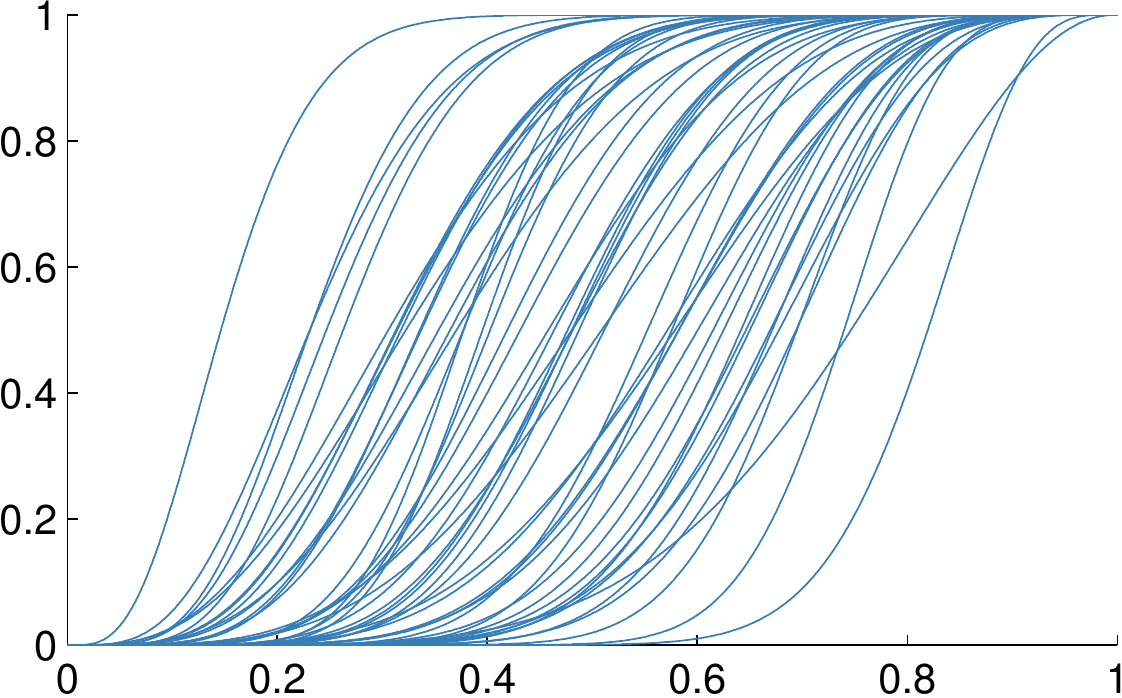}}
\caption{Each figure shows 50 warping functions resulting from the Beta CDF where the shape parameters $\alpha$ and $\beta$ are sampled from a log-normal prior with a different mean and variance.  The flexible Beta CDF captures many desirable warping functions and adjusting the prior over input warpings allows one to easily encode prior beliefs over the form of non-stationarity. For example, choosing $\mu_{\alpha}=\mu_{\beta}=0$ and $\sigma_{\alpha}=\sigma_{\beta}=0.5$ expresses a prior for slight or no warping~(\ref{fig:linear_beta}).  Setting $\mu_{\alpha}=0$, $\sigma_{\alpha}=0.25$ and $\mu_{\beta}=\sigma_{\beta}=1$ or $\mu_{\alpha}=\sigma_{\alpha}=1$, $\mu_{\beta}=0$ and $\sigma_{\beta}=0.25$ expresses a prior for approximately exponential~(\ref{fig:exp_beta}) or logarithmic~(\ref{fig:log_beta}) warping functions respectively. Approximately sigmoidal (\ref{fig:sigmoidal_beta}) warpings that contract the outer regions of the space while expanding the center can be expressed as a prior with $\mu_{\alpha}=\mu_{\beta}=2$ and $\sigma_{\alpha}=\sigma_{\beta}=0.5$.  One can also express logit shaped warpings (not shown here).} 
\label{fig:betadrawsfromprior}
\end{figure*}

\subsection{Integrating over warpings}
Rather than assume a single, explicit transformation function, we define a hierarchical Bayesian model by placing a prior over the shape parameters, $\alpha_d$ and $\beta_d$, of the bijections and integrating them out.  We treat the collection~$\{\alpha_d,\beta_d\}^D_{d=1}$ as hyperparameters of the covariance function and use Markov chain Monte Carlo via slice sampling, following the treatment of covariance hyperparameters from \citet{snoek-etal-2012b}.  We use a log-normal distribution, i.e. 
\begin{align}
\log(\alpha_d) &\sim \distNorm(\mu_{\alpha}, \sigma_{\alpha}) &
\log(\beta_d) &\sim \distNorm(\mu_{\beta}, \sigma_{\beta}),
\end{align}
to express a prior for a wide family of desirable functions.  Figure~\ref{fig:betadrawsfromprior} demonstrates example warping functions arising from sampling transformation parameters from various instantiations of this prior.  Note that the geometric mean or median of the zero-mean log-normal distribution for the $\alpha_d$ and $\beta_d$ corresponds to the identity transform.  With this prior the model centers itself on the identity transformation of the input space.  In the following empirical analysis we use this formulation with a variance of $0.75$.  A nice property of this approach is that a user can easily specify a prior when they expect a specific form of warping, as we show in~Figure~\ref{fig:betadrawsfromprior}.  

\subsection{Multi-Task Input Warping}
When training the same model on different datasets, certain properties, such as the size of the dataset, can have a dramatic effect on the optimal hyperparameter settings. For example, a model trained on a small dataset will likely require more regularization than the same model trained on a larger dataset. In other words, it is possible that one part of the input space on one task can be correlated with a different part of the input space on another task. To account for this, we allow each task to have its own set of warping parameters. Inferring these parameters will effectively try to warp both tasks into a jointly stationary space that is more suitably modeled by a standard multi-task kernel. In this way, large values on one task can map to small values on another, and vice versa.

\begin{table*}[t]
  \centering%
  \begin{tabular}[b]{lrrrr|rr}
    \toprule
    Experiment & \# Evals & \multicolumn{1}{c}{SMAC} & \multicolumn{1}{c}{Spearmint} & \multicolumn{1}{c|}{TPE} & \# Evals & \multicolumn{1}{c}{Spearmint + Warp}\\
    \cmidrule(r){1-7}
    Branin (0.398) & 200 & $0.655\pm 0.27$ & $\bf 0.398\pm 0.00$ & $0.526\pm 0.13$ & 40 & $\bf 0.398\pm 0.00$\\
    Hartmann 6 (-3.322) & 200 & $-2.977\pm 0.11$ & $-3.133 \pm 0.41$ & $-2.823\pm 0.18$ & 100 & $\bf -3.3166\pm 0.02$\\
    Logistic Regression & 100 & $8.6 \pm 0.9$ & $7.3\pm 0.2$ & $8.2\pm 0.6$ & 40 & $\bf 6.88 \pm 0.0$\\    
    LDA (On grid) & 50 & $1269.6\pm 2.9$ & $1272.6\pm 10.3$ & $1271.5\pm 3.5$ & 50 & $\bf 1266.2 \pm 0.1$ \\
    SVM (On grid) & 100 & $\bf 24.1\pm 0.1$ & $24.6\pm 0.9$ & $24.2\pm 0.0$ & 100 & $\bf 24.1 \pm 0.1$ \\
   \bottomrule
\end{tabular}
  \caption{We evaluate our algorithm on the continuous-valued parameter benchmarks proposed in \citet{Eggensperger-etal-2013a}. We compare to Sequential Model Based Algorithm Configuration (SMAC)~\citep{hutter-2011a}, the Tree Parzen Estimator (TPE)~\citep{BergstraJ2011} and Spearmint~\citep{snoek-etal-2012b}. The results for SMAC, Spearmint and TPE are reproduced from \citet{Eggensperger-etal-2013a}.  Following the standard protocol for these benchmarks, each algorithm was run ten times for the given number of evaluations, and the average validation loss and standard deviation are reported.  The algorithm with the lowest validation loss is shown in bold.  We note that on some of the benchmarks our algorithm converges to a solution in far fewer evaluations than the protocol allows.}
  \label{tab:result_comparison}
\end{table*}

\section{Empirical Analyses}
\label{sec:empirical}
Our empirical analysis is comprised of three distinct experiments. In the first experiment, we compare to the method of~\citet{snoek-etal-2012b} in order to demonstrate the effectiveness of input warping. In the second experiment, we compare to other hyperparameter optimization methods using a subset of the benchmark suite found in~\citet{Eggensperger-etal-2013a}. Finally, we show how our multi-task extension can further benefit this important setting.
\subsection{Comparison to Stationary GPs}
\paragraph{Experimental setup}
We evaluate the standard Gaussian process expected improvement algorithm (GP EI MCMC) as implemented by \citet{snoek-etal-2012b}, with and without warping.  Following their treatment, we use the Mat\'{e}rn~$5/2$ kernel and we marginalize over kernel parameters~$\theta$ using slice sampling~\citep{Murray-Adams-2010a}.  We repeat three of the experiments\footnote{See \citet{snoek-etal-2012b} for details of these experiments.} from~\citet{snoek-etal-2012b}, and perform an experiment involving the tuning of a deep convolutional neural network\footnote{We use the Deepnet package from \url{https://github.com/nitishsrivastava/deepnet}} on a subset of the popular CIFAR-10 data set~\citep{Krizhevsky-2009a}.  The deep network consists of three convolutional layers and two fully connected layers and we optimize over two learning rates, one for each layer type, six dropout regularization rates, six weight norm constraints, the number of hidden units per layer, a convolutional kernel size and a pooling size for a total of 21 hyperparameters.  On the logistic regression problem we also compare to warping the input space \emph{a priori} using the log-transform (optimizing in log-space). 

\begin{figure*}
\centering%
\subfloat[Logistic Regression]{%
  \includegraphics[width=0.24\textwidth]{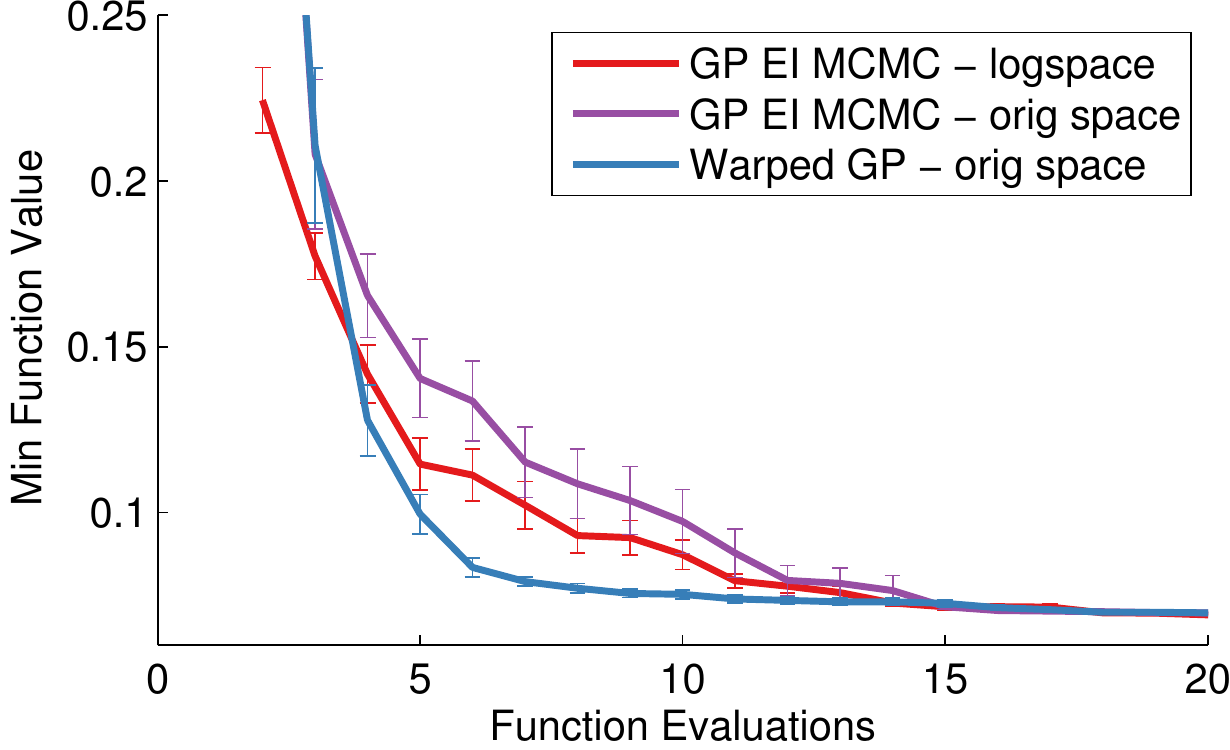}}
\subfloat[Online LDA\label{fig:lda_warped}]{%
  \includegraphics[width=0.24\textwidth]{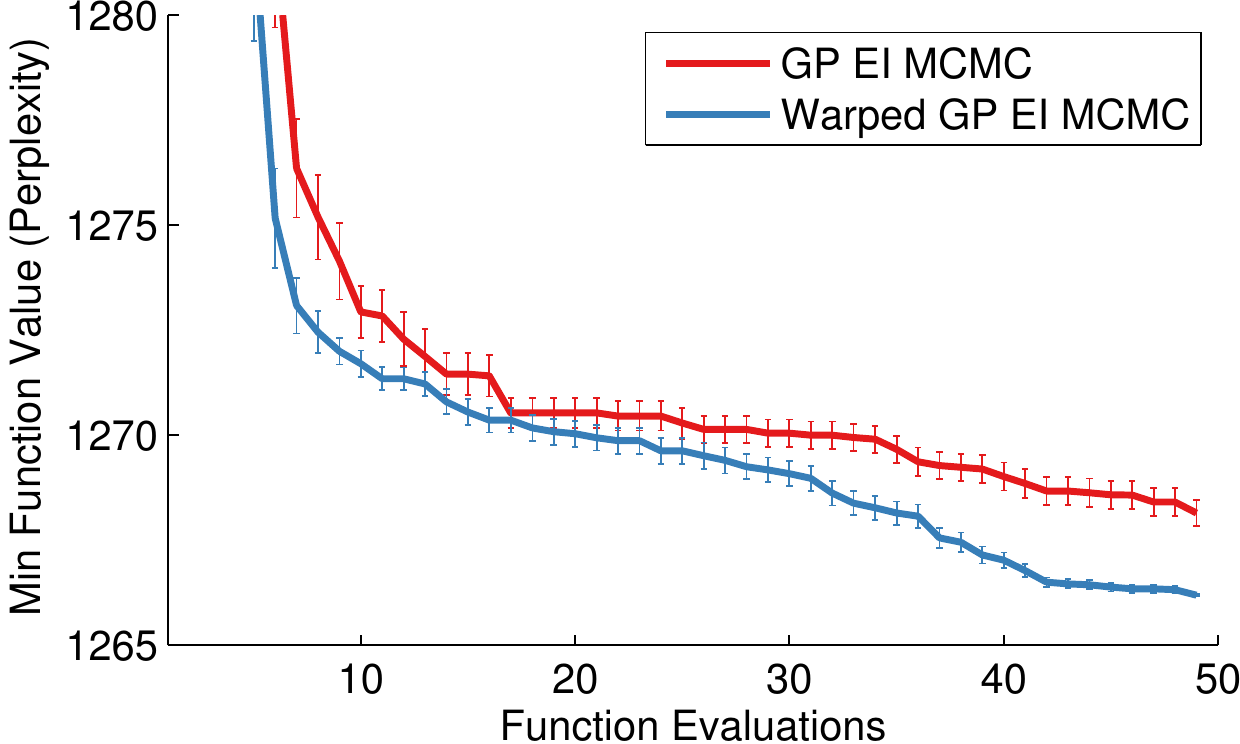}}
  \subfloat[Structured SVM\label{fig:ssvm_warped}]{%
  \includegraphics[width=0.24\textwidth]{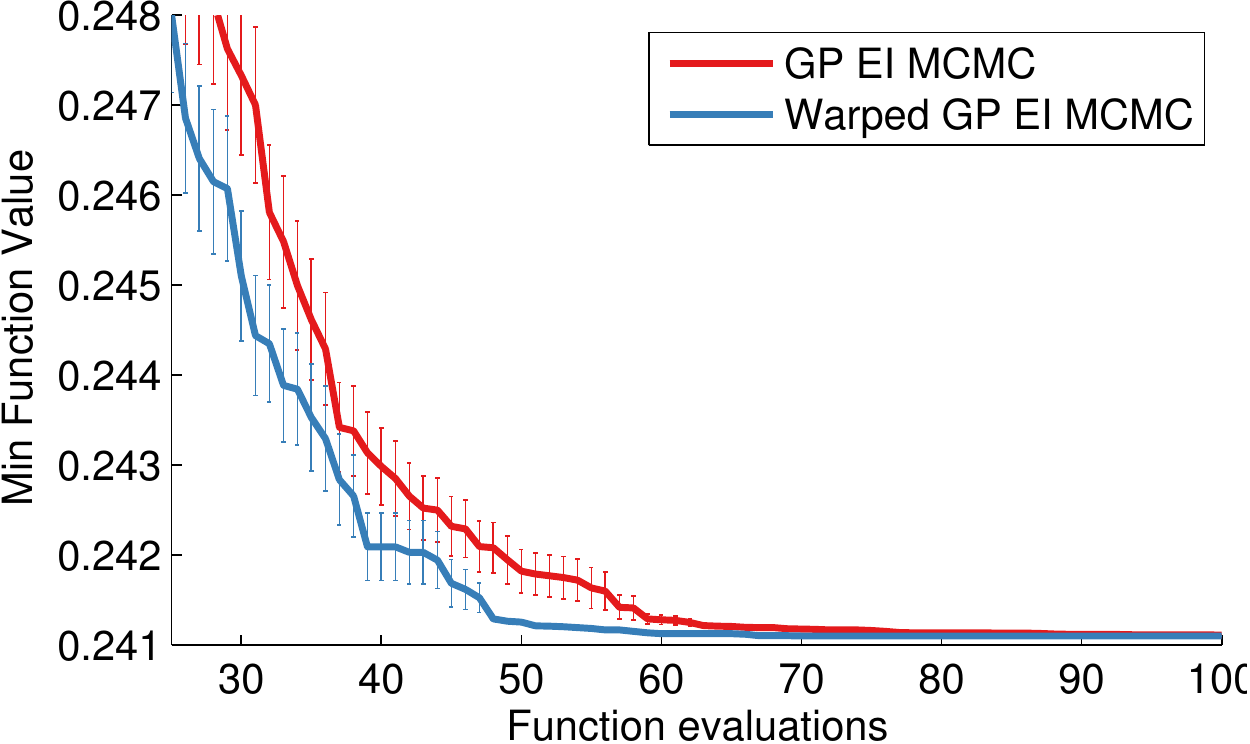}}
\subfloat[Cifar 10 Subset \label{fig:cifar_10_small}]{%
  \includegraphics[width=0.24\textwidth]{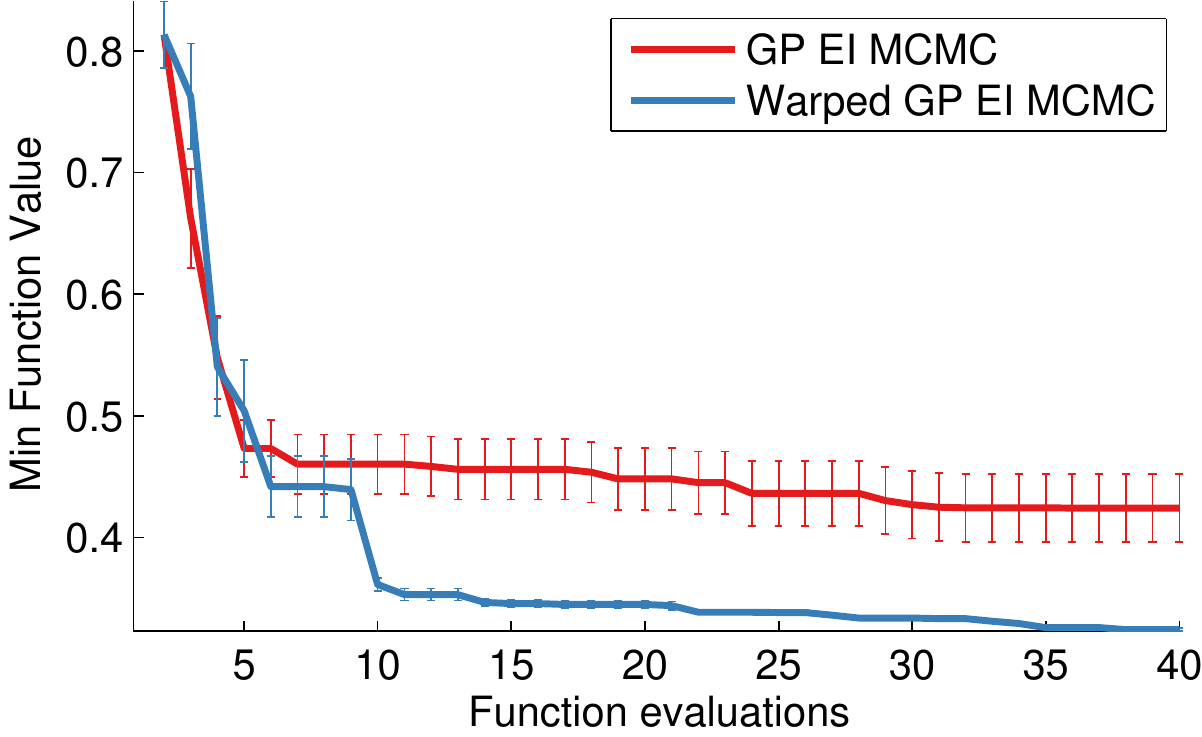}}
\caption{An empirical comparison of Bayesian optimization following the standard Gaussian process expected improvement algorithm (GP EI MCMC) and our strategy (Warped GP EI MCMC) for modeling input non-stationarity.  The methods are compared on four challenging problems involving the optimization of the hyperparameters of popular machine learning algorithms.}
\label{fig:warpingresults}
\end{figure*}

\paragraph{Results}
Figure~\ref{fig:warpingresults} shows that in all cases, dealing with non-stationary effects via input warpings greatly improves the convergence of the optimization. Of particular note, on the higher-dimensional convolutional network problem (Figure~\ref{fig:cifar_10_small}) input warped Bayesian optimization consistently converges to a better solution than Bayesian optimization with a stationary GP.

\begin{figure*}[t]
\centering%
  \subfloat[Logistic Regression\label{fig:logreg_warped}]{%
  \includegraphics[width=0.33\textwidth]{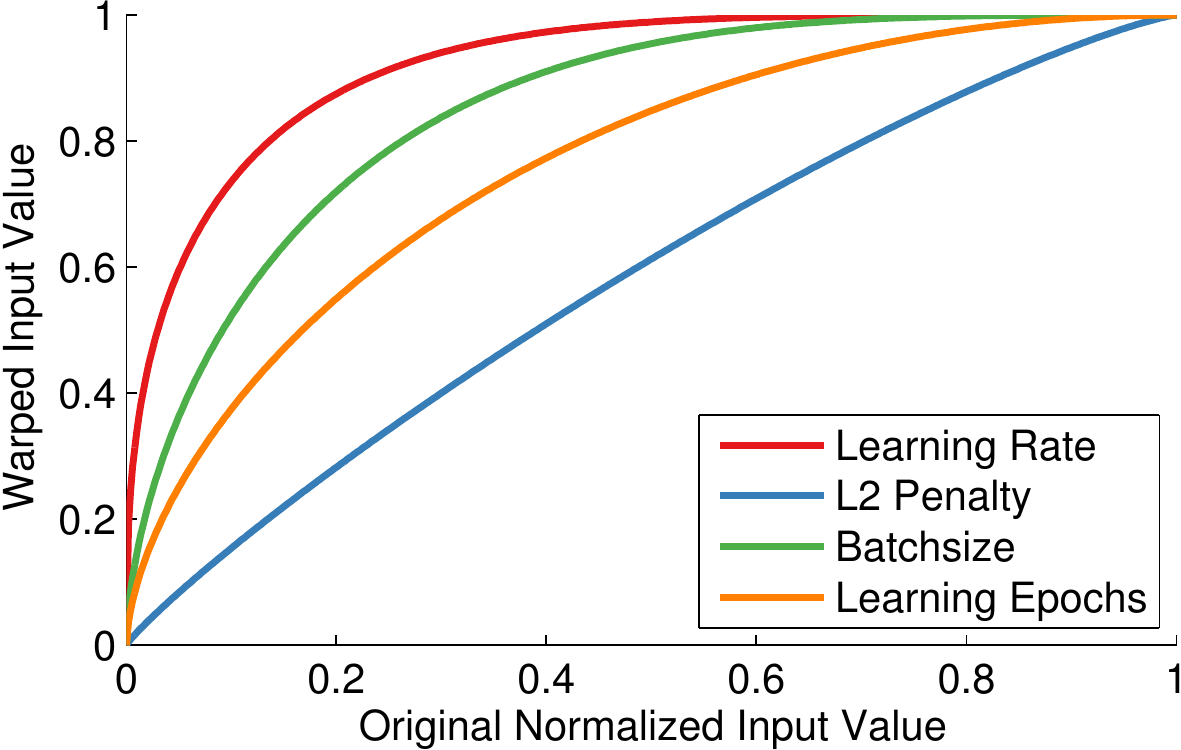}}
 \subfloat[Logistic Regression (Learning Rate)\label{fig:logreg_warped_lr}]{%
  \includegraphics[clip=true,trim=0 21 0 0,width=0.35\textwidth]{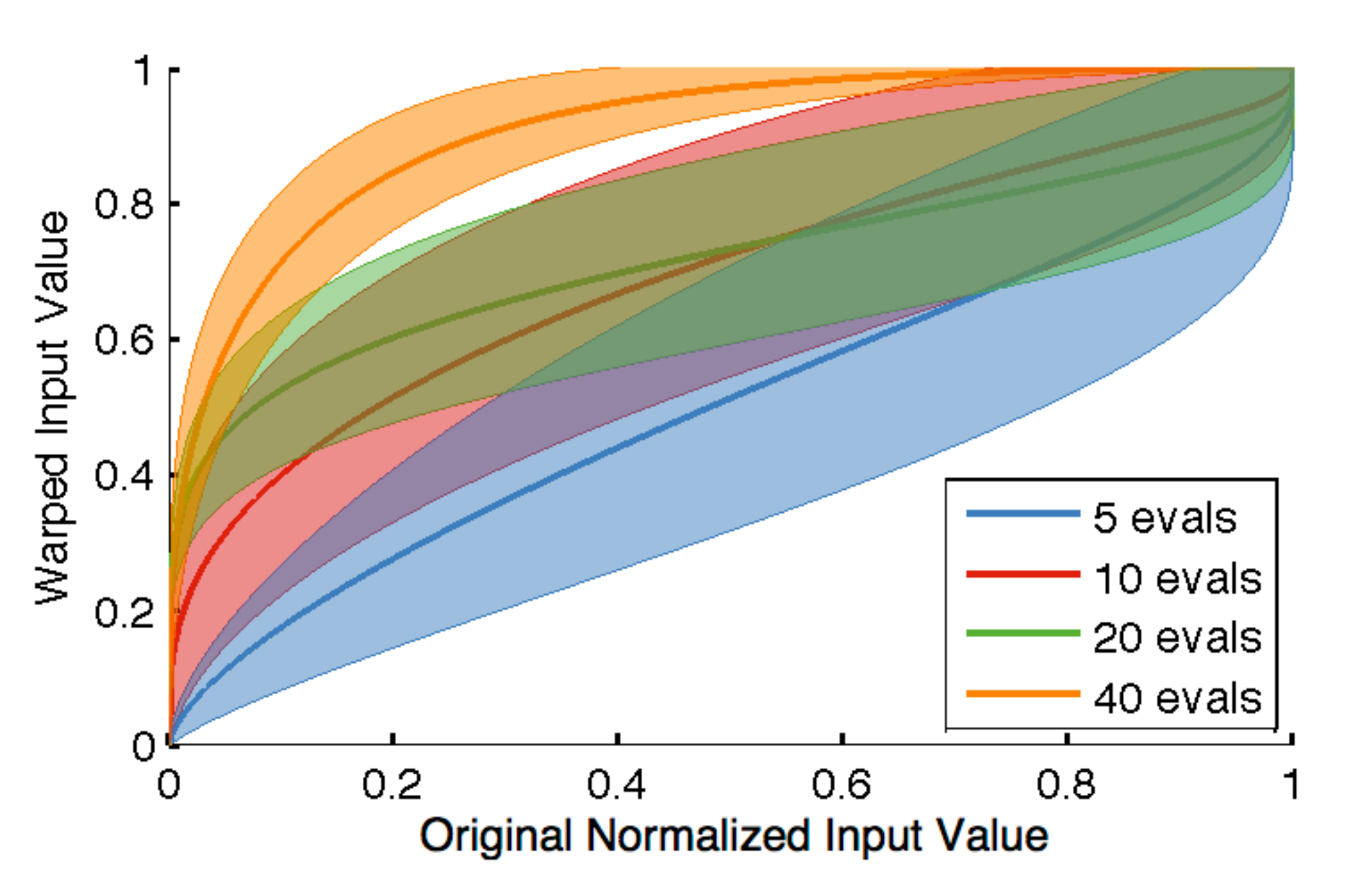}} \\
  \subfloat[DNN Learning Rates\label{fig:learnrates}]{%
  \includegraphics[width=0.33\textwidth]{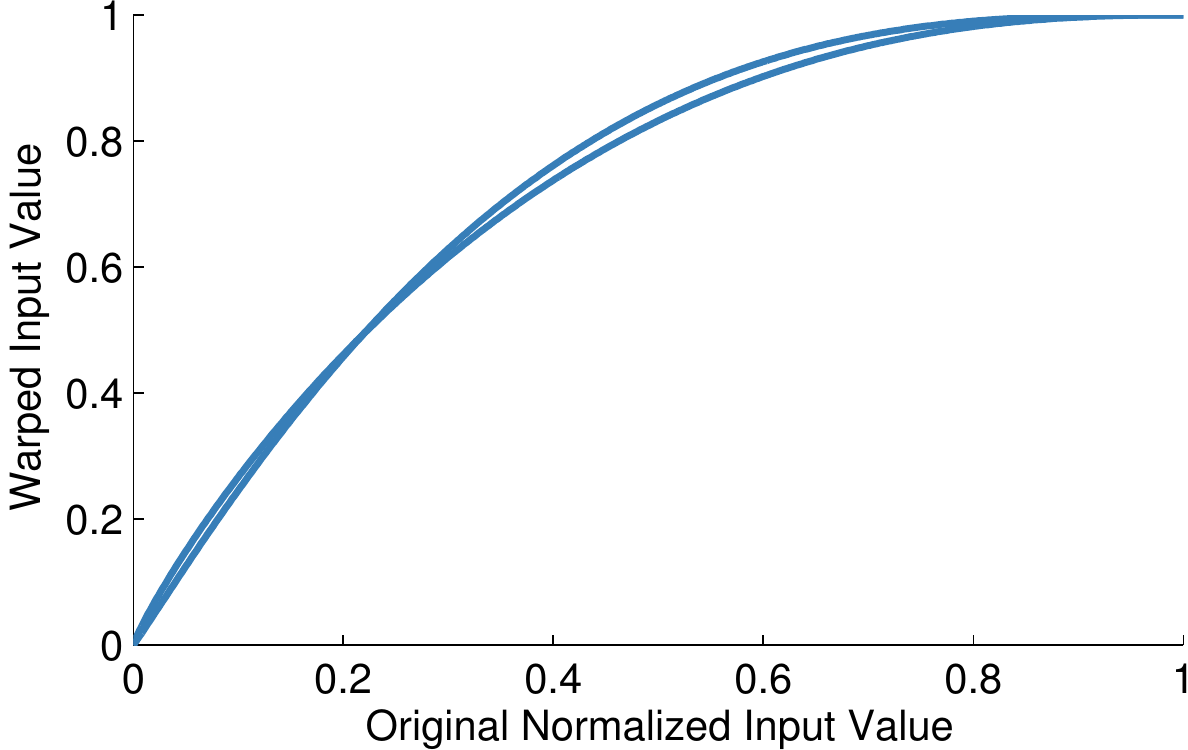}}
  \subfloat[DNN Weight Norms\label{fig:weightnorms}]{%
  \includegraphics[width=0.33\textwidth]{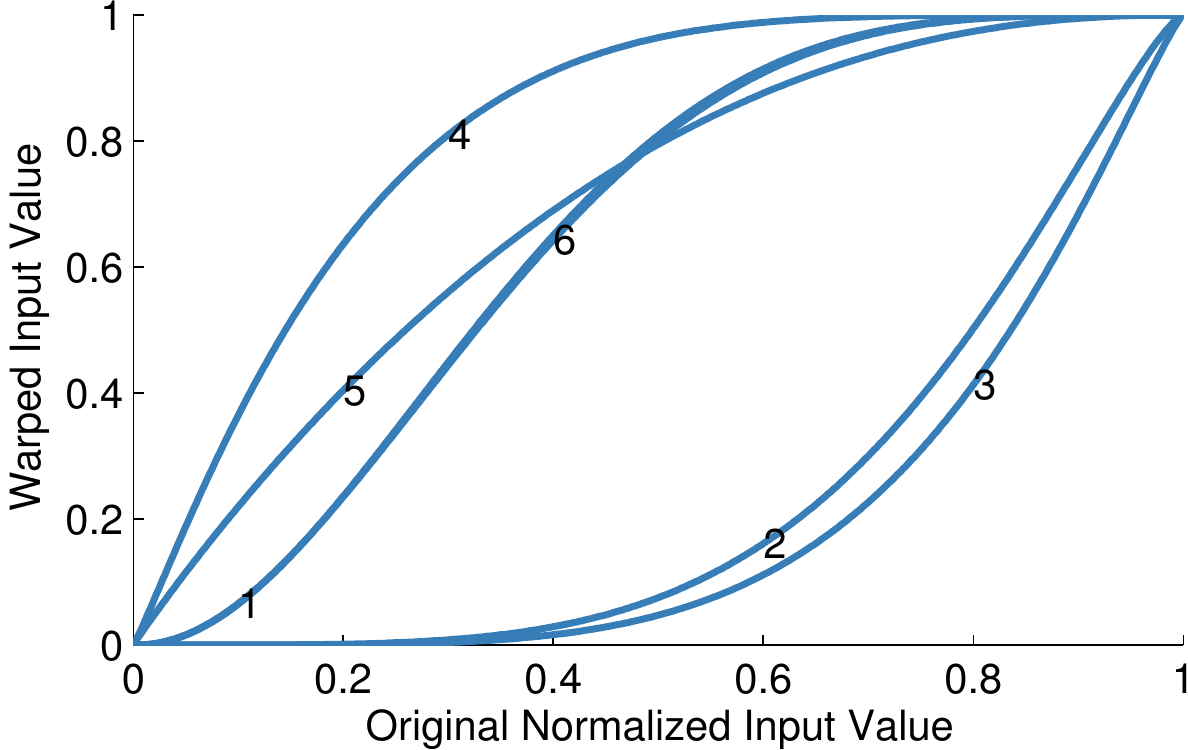}}
  \subfloat[DNN Dropout\label{fig:dropouts}]{%
  \includegraphics[width=0.33\textwidth]{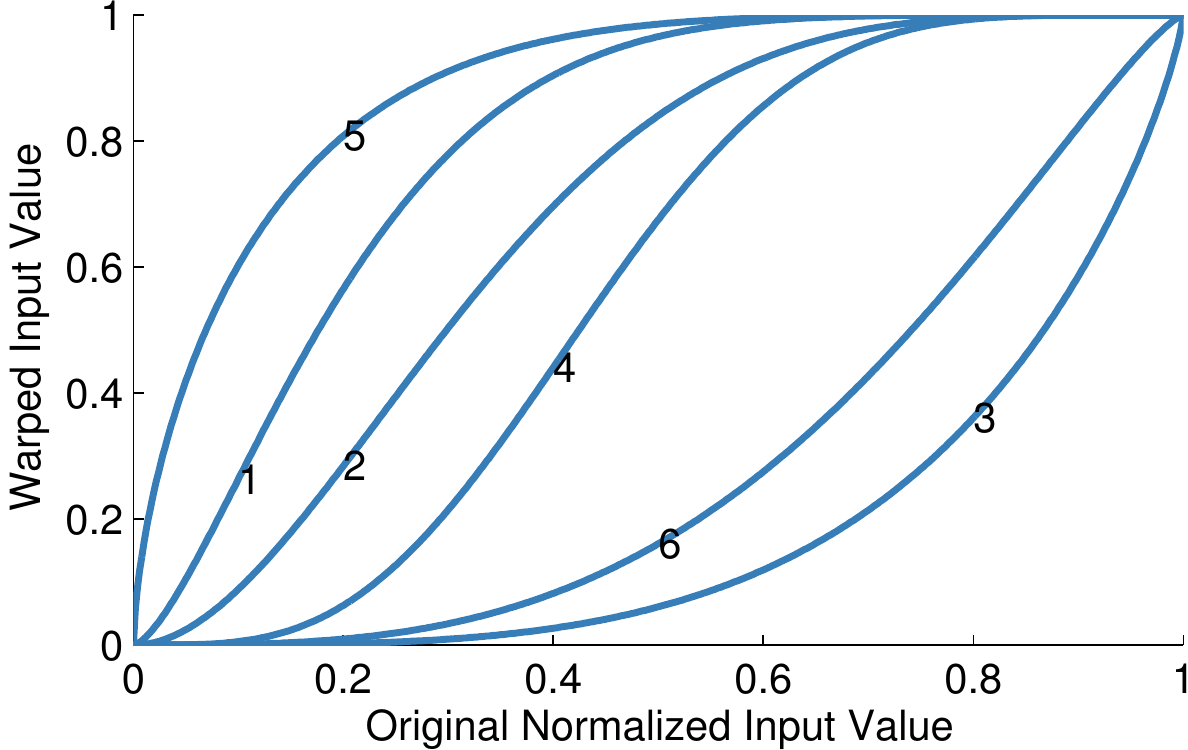}}
\caption{Example input warpings learned for the logistic regression problem (Figures~\ref{fig:logreg_warped},\ref{fig:logreg_warped_lr}) and the parameters of the deep convolutional neural network (Figures~\ref{fig:learnrates}, \ref{fig:dropouts}, \ref{fig:weightnorms}).  Each plot shows the mean warping, averaged over 100 samples, of each of the parameters.  Figure \ref{fig:logreg_warped_lr} shows the warping learned on the learning rate parameter for logistic regression with different numbers of observations, along with the standard deviation.  Each curve in Figures~\ref{fig:weightnorms} and \ref{fig:dropouts} is annotated with the depth of the layer that each parameter is applied to.}
\label{fig:learnedwarpings}
\end{figure*}

In Figure~\ref{fig:learnedwarpings} we plot examples of some of the inferred warpings. For logistic regression, Figure~\ref{fig:logreg_warped} shows that our method learns different logarithmic-like warpings for three dimensions and no warping for the fourth. Figure~\ref{fig:logreg_warped_lr} shows how the posterior distribution over the learning rate warping evolves, becoming more extreme and more certain, as observations are gathered. Figure~\ref{fig:learnrates} shows that on both convolutional and dense layers, the intuition that one should log-transform the learning rates holds. For transformations on weight norm constraints, shown in Figure~\ref{fig:weightnorms}, the weights connected to the inputs and outputs use a sigmoidal transformation, the convolutional-layer weights use an exponential transformation, and the dense-layer weights use a logarithmic transformation. Effectively, this means that the most variation in the error occurs in the medium, high and low scales respectively for these types of weights. Especially interesting are the wide variety of transformations that are learned for dropout on different layers, shown in Figure~\ref{fig:dropouts}. These show that different layers benefit from different dropout rates, which was also confirmed on test set error, and challenges the notion that they should just be set to $0.5$~\citep{hinton2012improving}.

It is clear that the learned warpings are non-trivial. In some cases, like with learning rates, they agree with intuition, while for others like dropout they yield surprising results. Given the number of hyperparameters and the variety of transformations, it is highly unlikely that even experts would be able to determine the whole set of appropriate warpings. This highlights the utility of learning them automatically.

\subsection{HPOLib Continuous Benchmarks}
\paragraph{Experimental setup}
In our next set of experiments, we tested our approach on the subset of benchmarks over continuous inputs from the HPOLib benchmark suite~\citep{Eggensperger-etal-2013a}. These benchmarks are designed to assess the strengths and weaknesses of several popular hyperparameter optimization schemes. All of the tested methods perform Bayesian optimization, however the underlying surrogate models differ significantly. The SMAC package~\citep{hutter-2011a} uses a random forest, the Hyperopt package~\citep{BergstraJ2011} uses the tree Parzen estimator, and the Spearmint package~\citep{snoek-etal-2012b} uses a Gaussian process. For our experiments, we augmented the Spearmint package with input warping.

\paragraph{Results}
Table~\ref{tab:result_comparison} shows the results, where all but the warped results are taken from~\citet{Eggensperger-etal-2013a}. Overall, input warpings improve the performance of the Gaussian process approach such that it does at least as well as every other method, and in many cases better. Furthermore, the standard deviation also decreases significantly in many instances, meaning that the results are far more reliable. Finally, it is worth noting that the number of function evaluations required to solve the problems is also drastically reduced in many cases.

Interestingly, the random forest approach in SMAC also naturally deals with nonstationarity, albeit in a fundamentally different way, by partitioning the space in a non-uniform manner. There are several possibilities to explain the performance discrepancy. Unlike random forests, Gaussian processes produce a smooth function of the inputs, meaning that EI can be locally optimized via gradient methods, so it is possible that better query points are selected in this way. Alternatively, the random forest is not a well-defined prior on functions and there may be overfitting in the absence of parameter marginalization. Further investigation is merited to tease apart this discrepancy.

\subsection{Multi-Task Warping}
\paragraph{Experimental setup}
In this experiment, we apply multi-task warping to logistic regression and online LDA~\citep{Hoffman2010} in a similar manner to \citet{swersky-etal-2013a}. In the logistic regression problem, a search over hyperparameters has already been completed on the USPS dataset, which consists of $6,000$ training examples of handwritten digits of size $16\times 16$. It was demonstrated that it was possible to use this previous search to speed up the hyperparameter search for logistic regression on the MNIST dataset, which consists of $60,000$ training examples of size $28\times 28$.

In the online LDA problem, we assume that a model has been trained on $50,000$ documents and that we would now like to train one on $200,000$ documents. Again, it was shown that it is possible to transfer information over to this task, resulting in more efficient optimization.

\begin{figure*}[t]
\centering%
  \subfloat[Logistic Regression Mean Warpings\label{fig:logreg_collab_warped_params}]{%
  \includegraphics[width=0.33\textwidth]{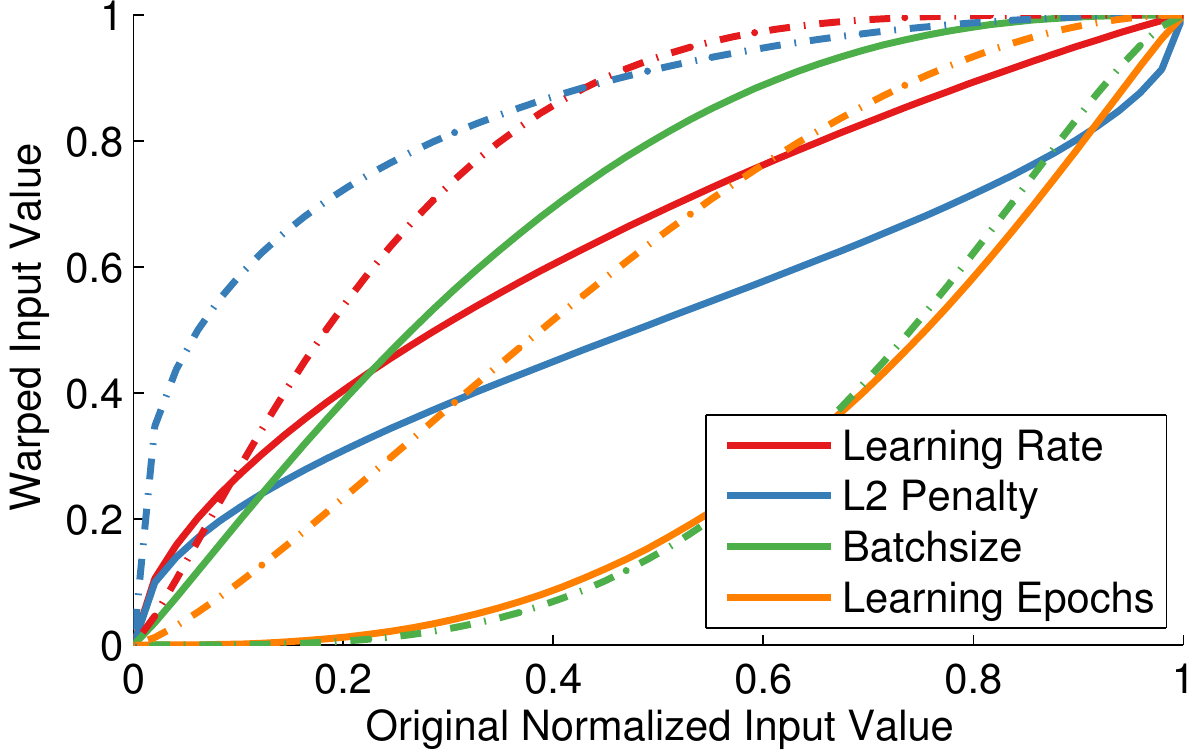}}
  \subfloat[Logistic Regression\label{fig:logreg_collab_warped}]{%
  \includegraphics[width=0.33\textwidth]{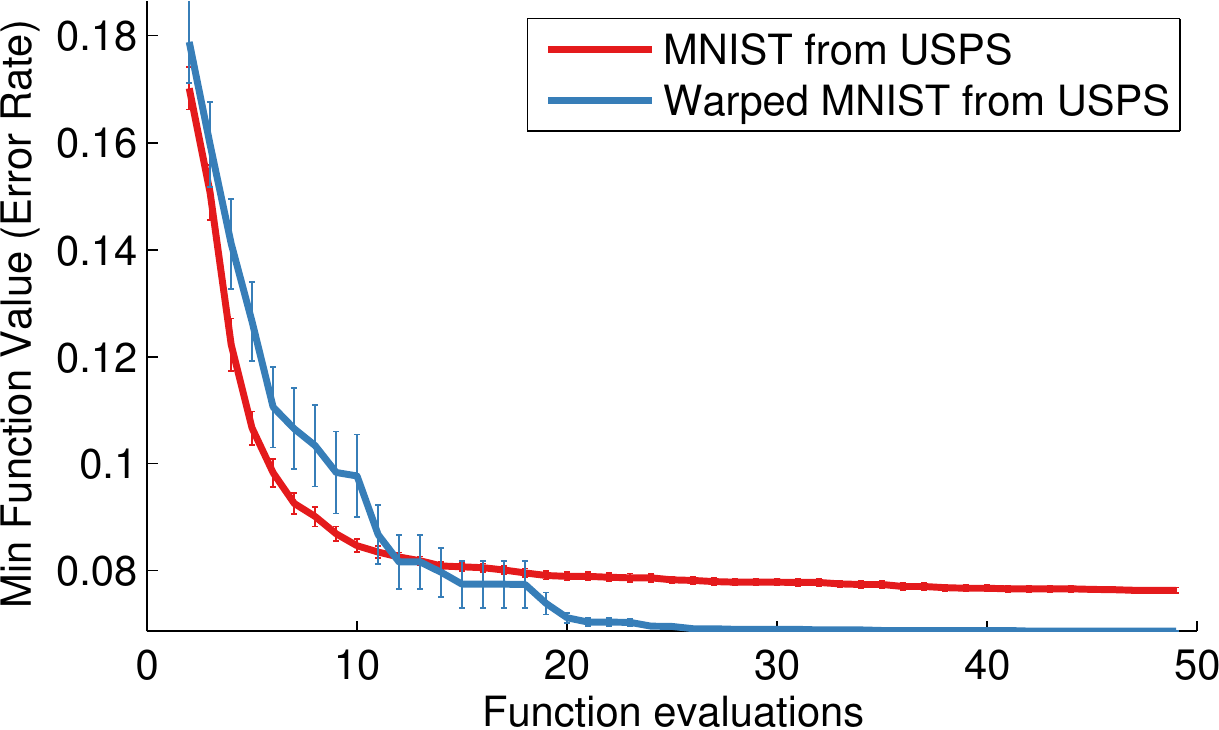}}
  \subfloat[Online LDA\label{fig:lda_collab_warped}]{%
  \includegraphics[width=0.33\textwidth]{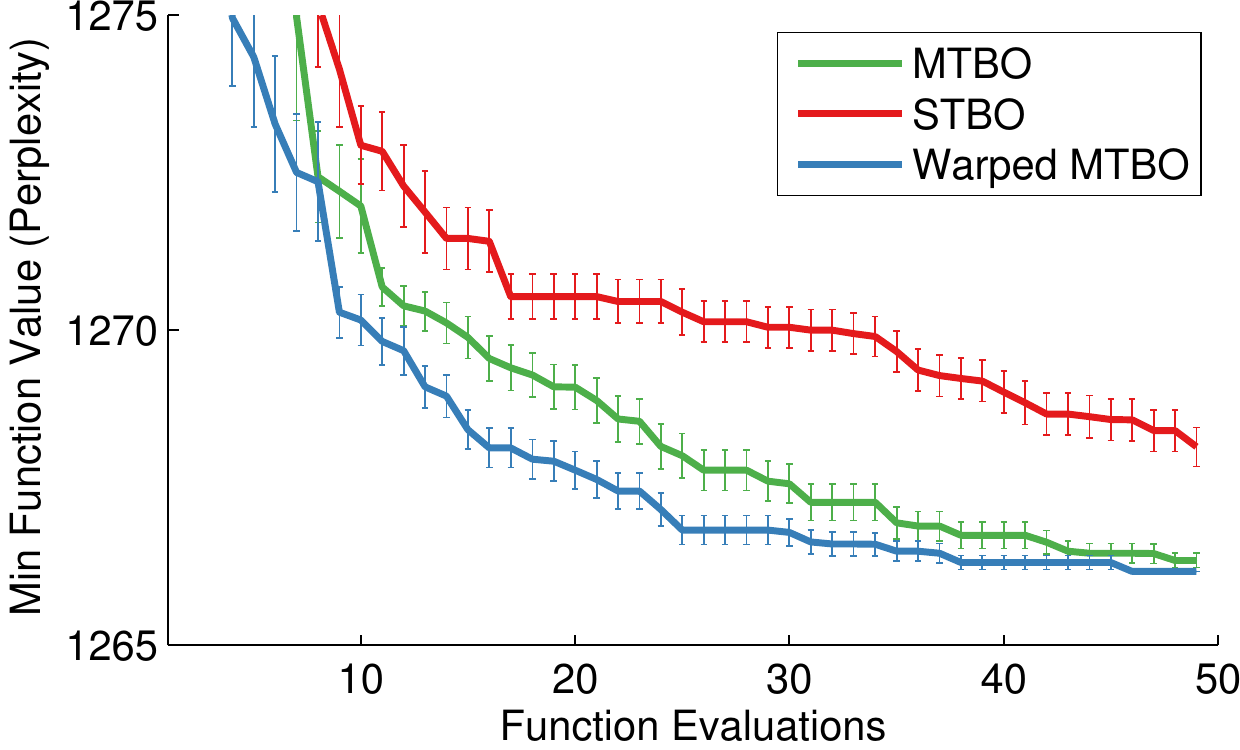}}

\caption{Multi-task warping applied to logistic regression and online LDA. For logistic regression, the model is trained on USPS first and then the search is transferred to MNIST. For online LDA, the data is subsampled and the model is learned, then the search transferred to the full dataset. In both cases, the warped version substantially outperforms multi-task Bayesian optimization with no warping. In Figure \ref{fig:logreg_collab_warped_params}, we show the mean warping learned by each task for each parameter. The solid lines indicates the MNIST task, while the dashed lines indicate the USPS task.}
\label{fig:multitask_warping}
\end{figure*}

\paragraph{Results}
In Figure \ref{fig:multitask_warping} we see that warped multi-task Bayesian optimization (warped MTBO) outperforms multi-task Bayesian optimization (MTBO) without warping, and performs far better than single-task Bayesian optimization (STBO) that does not have the benefit of a prior search. On logistic regression it appears that ordinary MTBO gets stuck in a local minimum, while warped MTBO is able to consistently escape this by the $20^\text{th}$ function evaluation.

In Figure \ref{fig:logreg_collab_warped_params} we show the mean warping learned for each task/hyperparameter combination (generated by averaging over samples from the posterior). The warping of the $\mathrm{L}_2$ penalty on the USPS model favours configurations that are toward the higher end of the range. Conversely, the warping on the MNIST dataset favours relatively lower penalties. This agrees with intuition that a high regularization with less data is roughly equivalent to low regularization with more data. Other observations also agree with intuition. For example, since USPS is smaller each learning epoch consists of fewer parameter updates. This can be offset by training for more epochs, using smaller minibatch sizes, or increasing the learning rate relative to the same model on MNIST.

\section{Conclusion}

In this paper we develop a novel formulation to elegantly model non-stationary functions using Gaussian processes that is especially well suited to Bayesian optimization.  Our approach uses the cumulative distribution function of the Beta distribution to warp the input space in order to remove the effects of mild input-dependent length scale variations.  This approach allows us to automatically infer a variety of warpings in a computationally efficient way.  In our empirical analysis we see that an inability to model non-stationary functions is a major weakness when using stationary kernels in the GP Bayesian optimization framework.  Our simple approach to learn the form of the non-stationarity significantly outperforms the standard Bayesian optimization routine of~\citet{snoek-etal-2012b} both in the number of evaluations it takes to converge and the value reached. As an additional bonus, the method finds good solutions more reliably. Our experiments on the continuous subset of the HPOLib benchmark~\citep{Eggensperger-etal-2013a} shows that input warping performs substantially better than state-of-the-art baselines on these problems.

A key advantage of our approach is that the learned transformations can be analyzed \emph{post hoc}, and our analysis of a convolutional neural network architecture leads to surprising insights that challenge established doctrine.  Post-training analysis is becoming a critical component of neural network development. For example, the winning Imagenet 2013 \citep{deng2009imagenet} submission~\citep{zeiler2013visualizing} used \emph{post hoc} analysis to correct for model defects. The development of interpretable Bayesian optimization strategies can provide a unique opportunity to facilitate this kind of interaction. An interesting follow-up would be to determine whether consistent patterns emerge across architectures, datasets and domains.

In Bayesian optimization, properly characterizing uncertainty is just as important as making predictions. GPs are ideally suited to this problem because they offer a good balance between modeling power and computational tractability. In many real world problems, however, the assumptions made by the Gaussian processes are often violated, nullifying many of their benefits. In light of this, many opt to use frequentist models instead, which offer minimax-type guarantees. Our emphasis in this work is to demonstrate that it is possible to stay within the Bayesian framework and thus enjoy its characterization of uncertainty, while still overcoming some of the limitations associated with the conventional GP approach. In future work we intend to experiment with more elaborate models of non-stationarity to see if these yield further improvements.

\section*{Acknowledgements}
The authors would like to thank Nitish Srivastava for providing help with the Deepnet package.  Jasper Snoek is a fellow in the Harvard Center for
Research on Computation and Society.  During his time at the University of Toronto, Jasper Snoek was supported by a grant from Google.  This work was funded by DARPA Young Faculty Award N66001-12-1-4219, an Amazon AWS in Research grant, the Natural Sciences and Engineering Research Council of Canada (NSERC) and the Canadian Institute for Advanced Research (CIFAR).

\small
\bibliographystyle{unsrtnat}
\bibliography{draft}

\end{document}